\documentclass[10pt,twocolumn,letterpaper]{article}

\usepackage{cvpr}              
\definecolor{cvprblue}{rgb}{0.21,0.49,0.74}
\usepackage{multibib}
\newcites{supp}{References (Supplementary)}
\usepackage[pagebackref,breaklinks,colorlinks,allcolors=cvprblue]{hyperref}
\usepackage{array}
\usepackage{tabularx}
\newcolumntype{Y}{>{\centering\arraybackslash}X}
\usepackage{colortbl}
\usepackage{graphicx}
\usepackage{wrapfig}
\definecolor{yellow}{rgb}{1, 1, 0.7}
\definecolor{orange}{rgb}{1, 0.85, 0.7}
\definecolor{red}{rgb}{1, 0.7, 0.7}
\definecolor{normalred}{rgb}{1, 0, 0}
\usepackage[table]{xcolor}
\definecolor{mygray}{gray}{0.9}
\usepackage{enumitem}
\usepackage[ruled]{algorithm2e} 
\usepackage{xcolor}
\usepackage{multirow}
\usepackage{makecell}
\usepackage{chapterbib}
\definecolor{mygray}{gray}{0.94}

\def\paperID{} 
\def\confName{CVPR}
\def\confYear{2026}

\title{GaussianGrow: Geometry-aware Gaussian Growing from 3D Point Clouds with Text Guidance}

\author{
Weiqi Zhang$^{1*}$, Junsheng Zhou$^{1*\dag}$, Haotian Geng$^{1}$, Kanle Shi$^{2}$, Shenkun Xu$^{2}$, Yi Fang$^{3}$, Yu-Shen Liu$^{1\dag}$\\
School of Software, Tsinghua University, Beijing, China$^1$\\
Kuaishou Technology, Beijing, China$^2$\\
CAIR and CIDSAI, NYU Abu Dhabi, UAE$^3$\\
{\tt\small \{zwq23, zhou-js24, genght24\}@mails.tsinghua.edu.cn}\\
{\tt\small \{shikanle, xushenkun\}@kuaishou.com} {\tt\small yfang@nyu.edu} {\tt\small liuyushen@tsinghua.edu.cn}\\
}

\begin{document}

\twocolumn[{%
\renewcommand\twocolumn[1][]{#1}%
\maketitle
\begin{center}
\centering
\captionsetup{type=figure}
\vspace{-8mm}
\includegraphics[width=1\linewidth]{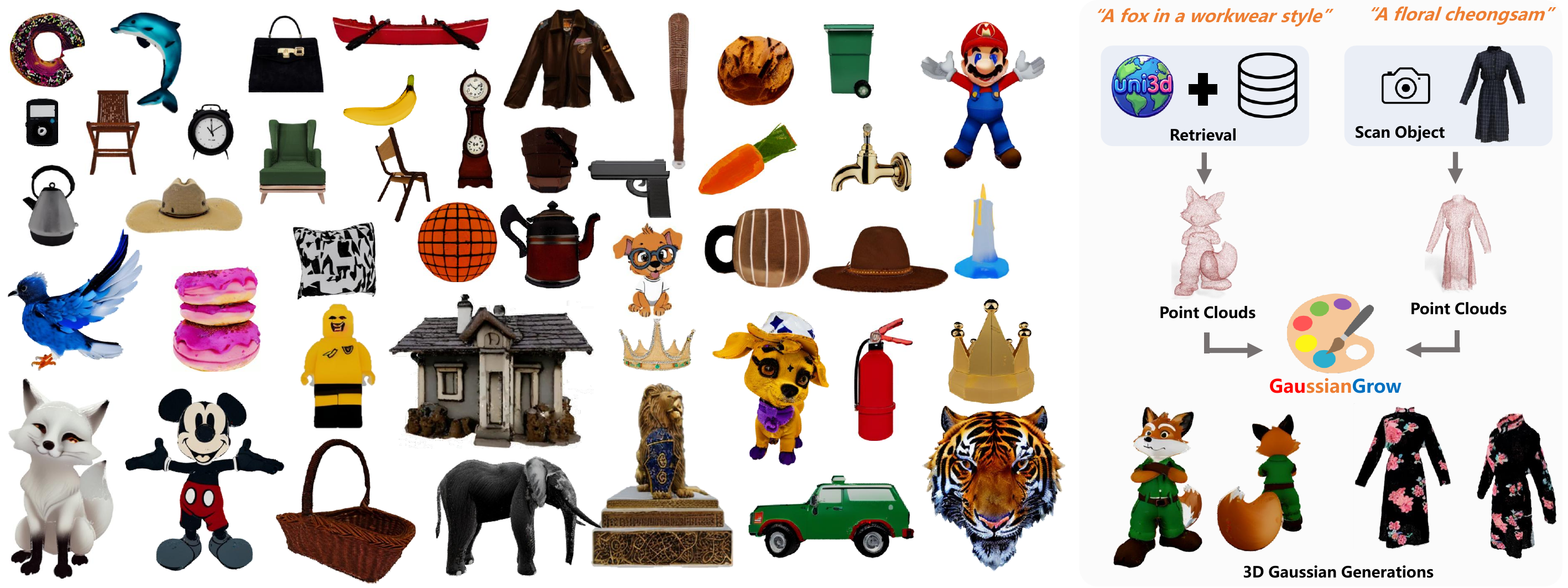}
\vspace{-8mm}
\captionof{figure}{\textbf{Left:} Diverse shapes generated by GaussianGrow. \textbf{Right:} The Gaussian generation pipeline of GaussianGrow. Reference point clouds can be obtained through large-scale retrieval or sensor scanning, from which Gaussians are grown under text guidance.
}
\label{fig:teaser}
\end{center}%
}]

\if TT\insert\footins{\footnotesize{
*Equal contribution. $\dag$ Corresponding authors.}}\fi

\begin{abstract}
3D Gaussian Splatting has demonstrated superior performance in rendering efficiency and quality, yet the generation of 3D Gaussians still remains a challenge without proper geometric priors. Existing methods have explored predicting point maps as geometric references for inferring Gaussian primitives, while the unreliable estimated geometries may lead to poor generations. In this work, we introduce GaussianGrow, a novel approach that generates 3D Gaussians by learning to grow them from easily accessible 3D point clouds, naturally enforcing geometric accuracy in Gaussian generation. Specifically, we design a text-guided Gaussian growing scheme that leverages a multi-view diffusion model to synthesize consistent appearances from input point clouds for supervision. To mitigate artifacts caused by fusing neighboring views, we constrain novel views generated at non-preset camera poses identified in overlapping regions across different views. For completing the hard-to-observe regions, we propose to iteratively detect the camera pose by observing the largest un-grown regions in point clouds and inpainting them by inpainting the rendered view with a pretrained 2D diffusion model. The process continues until complete Gaussians are generated. We extensively evaluate GaussianGrow on text-guided Gaussian generation from synthetic and even real-scanned point clouds. Project Page: \url{https://weiqi-zhang.github.io/GaussianGrow}.
\end{abstract}
\section{Introduction}
Creating high-quality 3D contents is a foundational task in 3D computer vision, driving various downstream applications such as film production, game design and AR/VR. 3D Gaussian Splatting (3DGS)~\cite{kerbl20233dgs} has emerged as an advanced 3D representation which enables high-fidelity 3D modeling with real-time rendering. The recent methods \cite{zhou2024diffgs,lin2025diffsplat, he2024gvgen, zhang2024gaussiancube} have thus focused on generating 3D contents as Gaussians. However, their generation quality remains limited due to the lack of proper geometry priors. Advanced approaches \cite{zou2023triplane, lu2024large} attempt to predict point maps as geometric references for inferring Gaussian primitives, but the unreliability of these estimated geometries often results in suboptimal generation.
Another series of research \cite{yu2023texture,chen2023text2tex,zeng2024paint3d} explores creating 3D content by texturing 3D meshes. The meshes provide clean 3D geometries, facilitating the learning of robust and accurate appearances. However, the reliance on 3D meshes as inputs demands extensive manual effort in 3D modeling. Additionally, their dependence on UV unwrapping introduces additional issues such as texture overlapping and distortion. While texture generation from meshes has been extensively studied, few approaches have explored modeling 3D point clouds for appearance generation.

With recent advances in 3D scanning devices such as LiDAR and depth cameras, collecting clean point cloud data has become significantly more convenient. In this paper, we demonstrate that the easily accessible 3D point clouds serve as a more flexible and reliable 3D geometric prior for improving 3D generation quality. Bridging the gap between point cloud geometries and 3D Gaussian Splatting appearances, we introduce a novel perspective that rethinks Gaussian generation by growing 3D Gaussians directly from point clouds. We propose GaussianGrow, a text-guided Gaussian generation framework that derives supervision for Gaussian optimization from synthesized multi-view images generated by a multi-view diffusion model. To further leverage the geometric prior, we optimize an unsigned distance field from the point cloud in an unsupervised manner.  We ensure view consistency by guiding the multi-view diffusion model with the robust geometric cues from point clouds and the distance field which produces robust depth and normal maps.

The overlapping regions across different generated views often cause artifacts due to challenges in fusing Gaussian primitives. To address this issue, we identify the camera poses that best observe these overlapping regions, generate additional appearances at those poses, and introduce constraints for overlap fusion. The pose identification is implemented as a camera pose optimization process that maximizes the parallelism with normals at visible points in the overlapping regions. After Gaussian growing and overlap refinement from the generated multi-views, some hard-to-observe regions still require further completion. We propose an iterative Gaussian inpainting scheme that gradually grows Gaussians in invisible point cloud regions. Specifically, we introduce an optimization-based approach that iteratively predicts the camera pose observing the largest un-grown regions. From this pose, we render an occluded view and utilize a pretrained 2D diffusion model for image inpainting, where the inpainted view serves as guidance for growing Gaussians in previously invisible regions. The Gaussian inpainting process iterates until complete Gaussians are generated.

We extensively evaluate GaussianGrow under the point-to-Gaussian task across diverse datasets, including both synthetic and real-world scanned point cloud models. Furthermore, we implement a text-to-3D generation pipeline by first retrieving point clouds using the multi-modal model Uni3D \cite{zhou2023uni3d}, then generating 3D Gaussians from them with GaussianGrow. Comprehensive experiments demonstrate our non-trivial improvements over the state-of-the-art methods. Our contributions can be summarized as follows:

\begin{itemize}
\item We propose GaussianGrow, a novel approach that generates 3D Gaussians by learning to grow them from easily accessible 3D point clouds with supervisions from multi-view diffusion models. By bridging the gap between point cloud geometries and 3D Gaussian Splatting appearances, GaussianGrow naturally enforces geometric accuracy in Gaussian generation. 
\item We introduce a refinement operation specifically designed to generate more consistent appearances in overlapping regions across neighboring views. An iterative Gaussian inpainting scheme is also proposed to first detect the camera pose by observing the largest invisible regions, then generate Gaussians on those un-grown points.
\item GaussianGrow achieves remarkable performance across various tasks, including point-to-Gaussian generation for both synthetic and real-scanned models, as well as text-to-3D generation.
\end{itemize}

\label{sec:intro}

\section{Related Work}
\label{sec:related}
\subsection{3D Generative Models}

Generative 3D modeling with neural networks has garnered growing interest in recent years. A prominent research direction \cite{sun2023dreamcraft3d,xu2023dream3d,poole2022dreamfusion,lin2023magic3d,wang2024prolificdreamer,raj2023dreambooth3d,metzer2023latent,ma2023geodream} focuses on optimization-based approaches, particularly Score Distillation Sampling (SDS), to synthesize 3D geometries by distilling knowledge from pretrained 2D diffusion models \cite{nichol2021improved, ho2020denoising}. While these methods produce compelling results, their reliance on per-instance optimization incurs substantial computational costs, which limits their scalability. Furthermore, SDS-based approaches often suffer from the Janus problem due to the absence of geometric priors.  Recent efforts \cite{tang2023volumediffusion,muller2023diffrf, wang2023rodin,zhou2024udiff} have explored direct generative modeling of 3D objects using diffusion models trained on 3D datasets. These models commonly represent radiance fields using triplanes \cite{wang2023rodin,shue20233d} or voxel grids \cite{muller2023diffrf,cheng2023sdfusion}.

With the advent of 3D Gaussian Splatting (3DGS) \cite{kerbl20233dgs}, generative modeling for 3DGS has emerged as a compelling yet challenging research direction. Unlike structured representations such as voxels or meshes, 3DGS exhibits irregular spatial distributions, making previous generative models designed for structured data unsuitable for direct application to 3DGS generation. Prior studies \cite{xu2024grm,zou2023triplane,hong2023lrm} have primarily focused on image-to-3D reconstruction, which relies on the quality and viewpoint of input images. 
Voxel-based approaches \cite{he2024gvgen,zhang2024gaussiancube} map Gaussians onto structured voxel grids, using volume-based networks for generative modeling. DiffGS \cite{zhou2024diffgs} represents 3DGS with several Gaussian functions to facilitate appearance generation, while DiffSplat \cite{lin2025diffsplat} repurposes image diffusion models for 3D Gaussian generation. However, all those methods follow the same Gaussian generation scheme to jointly learn both geometric structures and appearances, making them highly susceptible to failures when geometric predictions are inaccurate. Unlike MoRe~\cite{fang2026more}, which targets dynamic 4D reconstruction, GaussianGrow focuses on static 3D Gaussian generation by leveraging accessible point cloud geometries as reliable priors.


\subsection{Appearance Generation}

\noindent\textbf{Texture Generation for Meshes.} Generating high-quality textures for 3D meshes remains a significant challenge. Recent approaches leverage diffusion models trained on large-scale text-to-image datasets \cite{ho2020denoising} to enhance texture fidelity \cite{xiang2024make,cao2023texfusion,yu2023texture,liu2024text, jiang2024flexitex,xiong2024texgaussian,liu2024texoct,yu2024texgen,huo2024texgen}. Alternative methods \cite{richardson2023texture,chen2023text2tex,tang2024intex} employ depth-guided inpainting for viewpoint-consistent textures, while multi-view synthesis approaches \cite{zeng2024paint3d,bensadoun2024meta, cheng2024mvpaint} use geometric guidance to improve consistency. However, using 3D meshes as inputs requires significant manual effort in modeling. Moreover, their reliance on UV unwrapping introduces challenges such as texture overlap and distortion.

\noindent\textbf{Appearance Modeling for Point Clouds.} Recent studies have explored generating 3D Gaussian from colored point clouds \cite{lu2024large}, but reliance on RGB input limits applicability in color-scarce scenarios. GaussianPainter \cite{zhou2024gaussianpainter} enables stylization via reference images yet lacks fine-grained appearance control. These challenges underscore the need for an efficient framework that flexibly leverages the point cloud geometries for modeling high-quality 3D Gaussians. Recent works also explore material-aware Gaussian Splatting for physically based rendering~\cite{zhang2025materialrefgs}.
In this paper, we propose a novel perspective of Gaussian generation by learning to grow 3D Gaussian primitives from the point clouds, which can be easily accessed through scanning or cross-modal retrieval from existing point cloud databases. To this end, GaussianGrow generates high-quality 3D Gaussians with robust 3D geometric priors.


\begin{figure*}[!th]
    \centering
    \includegraphics[width=1\linewidth]{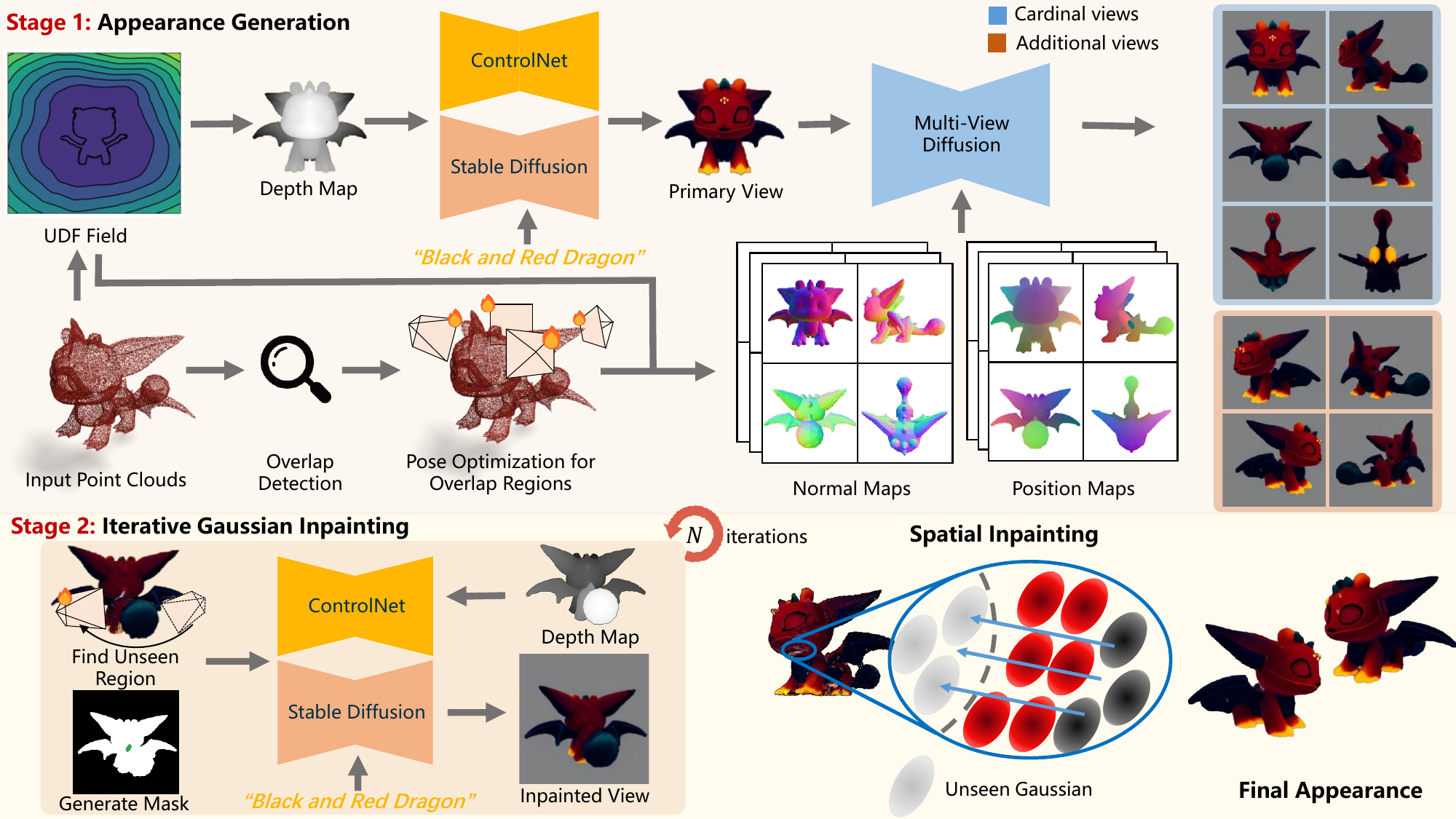}
    \vspace{-0.6cm}
    \caption{\textbf{Overview of GaussianGrow. Stage 1.} We leverage depth-aware ControlNet for primary view generation, with a geometry-aware diffusion model for multi-view synthesis. Additional views are generated for improving appearances in overlap regions by optimizing camera poses to observe overlap regions. Gaussians are optimized to grow with supervision from both cardinal and additional views. \textbf{Stage 2.} We iteratively inpaint Gaussians by optimizing camera poses to observe unseen regions, and inpaint them by inpainting the rendering view with a pretrained 2D diffusion model. The iteration continues until complete Gaussians are generated. A spatial Gaussian inpainting strategy is also used to diffuse appearance from optimized Gaussians to the hard-to-observe ones.} 
    \label{fig:method}
    \vspace{-0.6cm}
\end{figure*}

\section{Method}
\label{sec:method}
We present GaussianGrow, a novel generative model for 3D Gaussian Splatting by learning to grow 3D Gaussians from 3D point cloud geometries. Given a point cloud input $P=\{p_i\}_{i=1}^N$, GaussianGrow aims to learn high-fidelity  Gaussian primitives $G = \{g_i\}_{i=1}^M$, conditioned on a text prompt $c$. Fig.~\ref{fig:method} illustrates the overview of GaussianGrow.

\subsection{Preliminary Preparation}
\label{sec.3.1}
\noindent\textbf{Initialization Strategy.}
Proper initialization significantly improves the quality of Gaussian Generation. We initialize each Gaussian center at the corresponding point position in the input cloud $P=\{p_i\}_{i=1}^N$. To capture local geometry orientation, we optimize a neural Unsigned Distance Field (UDF) from $P$ using CAP-UDF~\cite{zhou2022capudf}, following recent progress in UDF-based geometry modeling~\cite{zhou2026udfstudio} Unlike Signed Distance Fields (SDFs)~\cite{nodamultipull,noda2025learning} that require watertight surfaces, UDFs can represent open and complex topologies, better suiting our setting. We compute normals $N=\{n_i\}_{i=1}^N$ through gradient prediction:
\begin{equation}
    n_i = \frac{\nabla f_u(p_i)}{\Vert \nabla f_u(p_i) \Vert}.
\end{equation}

In practice, we adopt 2D Gaussian Splatting representation~\cite{huang20242d} instead of vanilla 3DGS~\cite{kerbl20233dgs}, using oriented disks rather than ellipsoids to better represent detailed structures. Each Gaussian's rotation matrix $r_i$ is set according to its corresponding normal $n_i$, ensuring proper disk orientation for subsequent optimization.

\noindent\textbf{Geometric Information Maps.}
To extract comprehensive geometric information from the input point cloud, we compute three geometric representation maps: depth, normal, and position maps, each serving a distinct purpose in our pipeline. The depth map is essential for Depth-Aware-ControlNet~\cite{zhang2023adding}, guiding both primary view generation and image inpainting. We obtain the depth map $D_i$ through ray marching~\cite{Oechsle2021ICCV} in the learned unsigned distance field. While the normal and position maps are crucial for synthesizing consistent views with multi-view diffusion models. The normal map $N_i$ is obtained by inferring gradients at the zero-level set of the learned unsigned distance field. The position map $C_i$ maps each pixel to the exact XYZ coordinates of its corresponding point, providing robust spatial information. Note that our approach doesn't require any explicit mesh representations, where all the geometric information is derived from the point cloud and its distance field. 


\subsection{Appearance Generation}
\label{sec.3.2}

With the proper initialization, the next step is to generate appearances for Gaussian optimization. In practice, we adopt Hunyuan3D-Paint~\cite{hunyuan3d22025tencent} as the multi-view diffusion model for view synthesis. The commonly used setting for multi-view diffusion models is to generate six cardinal views. However, the overlapping areas across adjacent views in those pre-set camera poses often display inconsistencies in appearance.

\noindent\textbf{Overlap Detection.} 
To address the inevitable inconsistencies in the generations of the multi-view diffusion model, we propose a dense-view generation framework that extends beyond the standard six-view configuration for better appearance guidance at the overlap regions. 

\begin{figure}[ht!]
    \centering
    \vspace{-0.4cm}
    \includegraphics[width=1\linewidth]{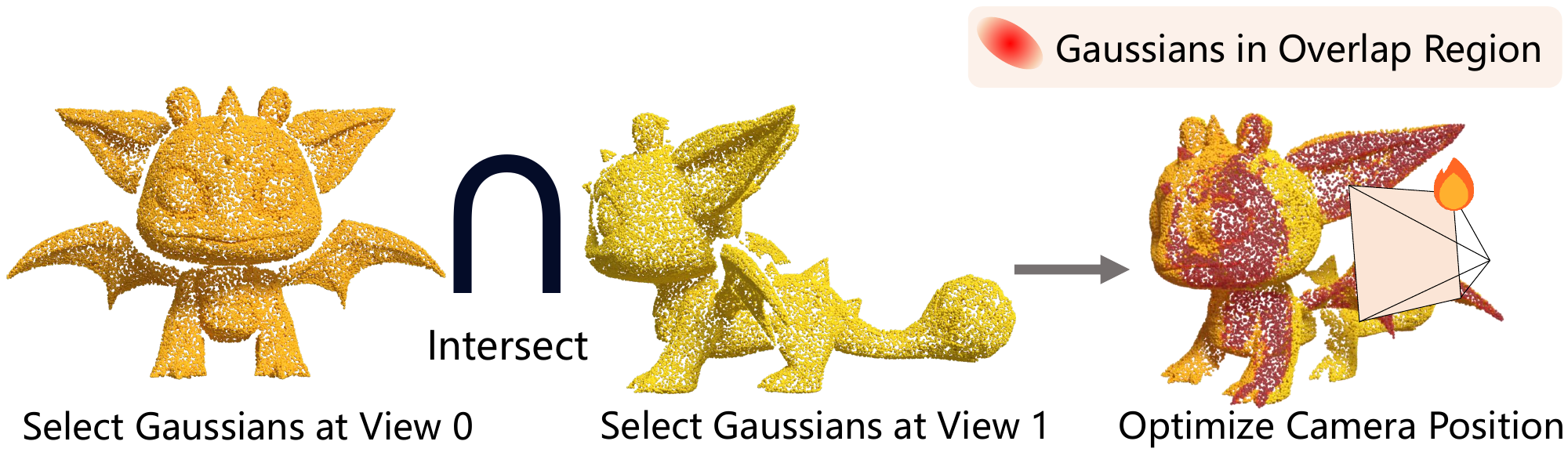}
    \vspace{-0.8cm}
    \caption{We obtain the additional camera poses by optimizing them to observe largest overlap regions.}
    \label{fig:inersect}
    \vspace{-0.4cm}
\end{figure}

Our method begins by identifying critical overlap regions where the inconsistencies are most pronounced. We first detect the visible Gaussians from each viewpoint $v_i$ by tracing rays from pixels and recording the first Gaussian intersected along each ray path. As illustrated in Fig.~\ref{fig:inersect}, for a pair of viewpoints $v_i$ and $v_j$ in standard cardinal views, we detect their overlap region $R_{i,j}$ by computing the intersection of their respective visible Gaussian sets. This approach ensures we focus only on those Gaussians that contribute to the appearance from both views simultaneously. We contribute a CUDA-based parallel implementation for speeding up this process, reducing computation time from minutes to seconds.

\noindent\textbf{Pose Optimization for Additional Views.} We then introduce additional camera poses with a special focus on refining overlap regions. To optimize these poses for enhanced appearance generation, we make them learnable through an optimization strategy that enforces alignment between the normal vectors of intersecting Gaussians and the corresponding camera rays. The key idea is that view directions most aligned with geometric normals capture the largest overlap regions, ultimately leading to greater consistency improvements.

Specifically, for each overlap region $R_{i,j}$, we define the loss as one minus the absolute cosine similarity between the normal $n_g$  at a Gaussian $g$ and the ray direction $d_{i,j}$ emitted from the camera at position $T_{i,j}$:
\begin{equation}
\mathcal{L}_{\text{align}} = \sum_{g \in R_{i,j}} \left( 1 - \left| \frac{\mathbf{d}_{i,j} \cdot \mathbf{n}_g}{\|\mathbf{d}_{i,j}\| \|\mathbf{n}_g\|} \right| \right).
\end{equation}

Minimizing $\mathcal{L}_{\text{align}}$ ensures that additional cameras are optimally positioned to align with local geometric structures, reducing projection distortions and enhancing view consistency. Note that the learnable camera poses are constrained to positions on a unit sphere centered at the object, with camera always pointing toward the center of the object.

\noindent\textbf{Multi-view Image Generation.} 
The next step is to generate high-quality appearances from both the pre-set six cardinal views and four additional views focusing on the main overlapping regions. To achieve this, We typically adopt an off-the-shelf SOTA multi-view diffusion model Hunyuan3D-Paint, which is trained with large-scale 3D/2D data, learning powerful geometry-aware view synthesis capabilities.

For each camera position, we render normal maps $N_i$ and position maps $C_i$ through the learned unsigned distance field. For the primary view, we additionally generate a depth map $D_i$ using ray marching on the field. This depth map, combined with the text prompt $c$, is processed through Depth-Aware-ControlNet~\cite{zhang2023adding} and Stable Diffusion~\cite{rombach2021highresolution} to produce a high-quality reference appearance $I_r$. The reference appearance serves as an anchor for maintaining consistent appearance across multiple viewpoints. Given the reference appearance and the geometric maps at all the $K=10$ views, the multi-view diffusion model finally generates high-fidelity appearance outputs $I = \{I_i\}_{i=1}^K$.

\noindent\textbf{Gaussian Optimization.} Our optimization strategy follows a two-phase approach that first addresses the six cardinal views $V =\{v_i\}_{i=1}^6$ before focusing on overlap regions. For each cardinal viewpoint $v_i$, we generate appearance images $I_i$ through the multi-view diffusion and optimize the corresponding visible Gaussians. Unlike previous approaches that optimize all Gaussians simultaneously across multiple views, we adapt a view-specific optimization scheme that focuses exclusively on the front-facing Gaussians visible from the current viewpoint without modifying the Gaussians on the back-facing surfaces. This targeted approach ensures that the well-optimized Gaussians are not affected at the optimization steps where they are not visible.

\begin{figure}[!t]
    \centering
    \includegraphics[width=1\linewidth]{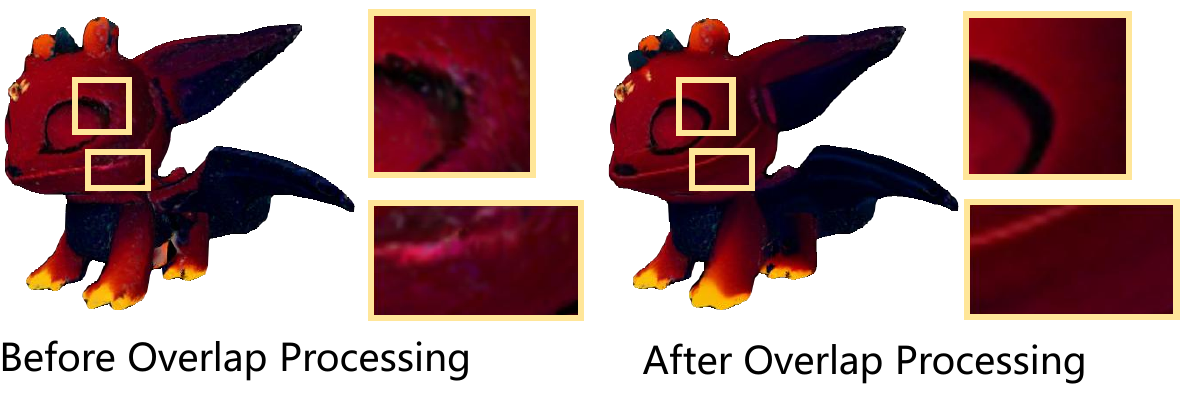}
    \vspace{-0.8cm}
    \caption{The effectiveness of processing overlap regions.}
    \label{fig:overlap}
    \vspace{-0.8cm}
\end{figure}

After optimizing the cardinal views, we proceed to refine the Gaussians in the detected overlap regions $R_{i,j}$ using our additional optimized camera positions $T_{i,j}$. This sequential approach ensures that the base appearance is well-established before addressing potential inconsistencies in transition areas. For each additional viewpoint $T_{i,j}$ targeting an overlap region $R_{i,j}$, we perform optimization only on the visible Gaussians $g_i \in G_{R_{i,j}}$ within $R_{i,j}$. Fig.~\ref{fig:overlap} demonstrates the effectiveness of our overlap region optimization in achieving seamless visual consistency.

\begin{figure}[!b]
    \centering
    \vspace{-0.7cm}
    \includegraphics[width=1\linewidth]{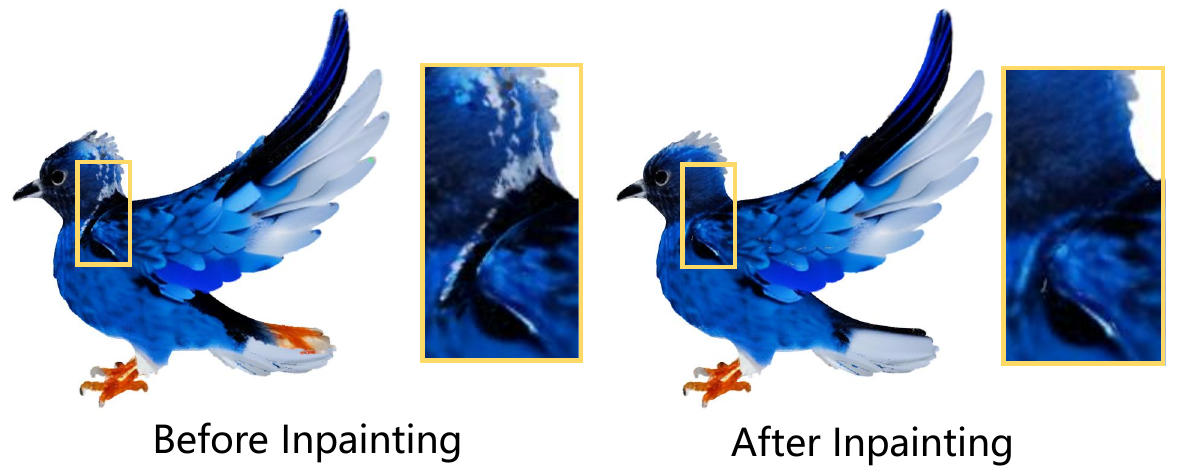}
    \vspace{-0.8cm}
    \caption{The effect of Gaussian inpainting.}
    \label{fig:ablation}
    \vspace{-0.8cm}
\end{figure}

\subsection{Iterative Inpainting and Refinement}
\label{sec.3.3}
\noindent\textbf{Find Unseen Region.}
Despite the multi-view appearance generation process covering most of the object's surfaces, certain regions may remain unseen or inadequately captured. Due to the diverse geometric structures of different objects, using a fixed set of dense viewpoints struggles to cover completely all parts of an object. Therefore, we need an automated approach to identify these unseen regions and optimize camera positions accordingly. 

To systematically identify the unseen regions, we propose a visibility-based optimization approach that predicts camera poses observing the largest invisible regions in the point cloud. Same to the pose optimization for additional views, we constrain the camera to positions on a unit sphere centered at the object. The core idea of the optimization is to analyze occlusion patterns among Gaussians by projecting them onto a 2D image plane from the current viewpoint and evaluating their depth relationships. The key insight is that camera poses where fewer unoptimized points are occluded by optimized ones, and where unoptimized points are positioned closer to the camera than optimized ones, enable better visibility of unseen regions.
To this end, we specifically quantify the number of unoptimized Gaussians occluded by optimized Gaussians as the optimization target. An occlusion is identified when the projections of two Gaussians overlap on the image plane according to their projected radii in 2D, with one Gaussian positioned in front of the other from the camera's perspective.


For each pair of Gaussians $(g_i,g_j)$, where $g_i$ is unoptimized and $g_j$ is optimized, we formulate the occlusion loss as:
{\small
\begin{equation}
    \mathcal{L}_{\text{occ}} = 
    \sum_{i, j}
    \sigma\Bigl((\tau (\rho_i + \rho_j)^2 - \|q_i - q_j\|^2)\Bigr)
    \sigma\Bigl(\tau(z_i - z_j)\Bigr),
\end{equation}
}
where $q_i$ and $q_j$ are the 2D projections of Gaussians on the image plane, $\rho_i$ and $\rho_j$ are their respective projected radii, $z_i$ and $z_j$ are their depths from the camera, $\sigma$ is the sigmoid function, and $\tau$ is a temperature parameter that controls the sharpness of the comparison.  This differentiable formulation enables efficient gradient descent optimization of learnable camera poses, enabling the systematic discovery of viewpoints that reveal largest unseen regions of the object.


\begin{figure*}
    \centering
    \vspace{-0.8cm}
    \includegraphics[width=1\linewidth]{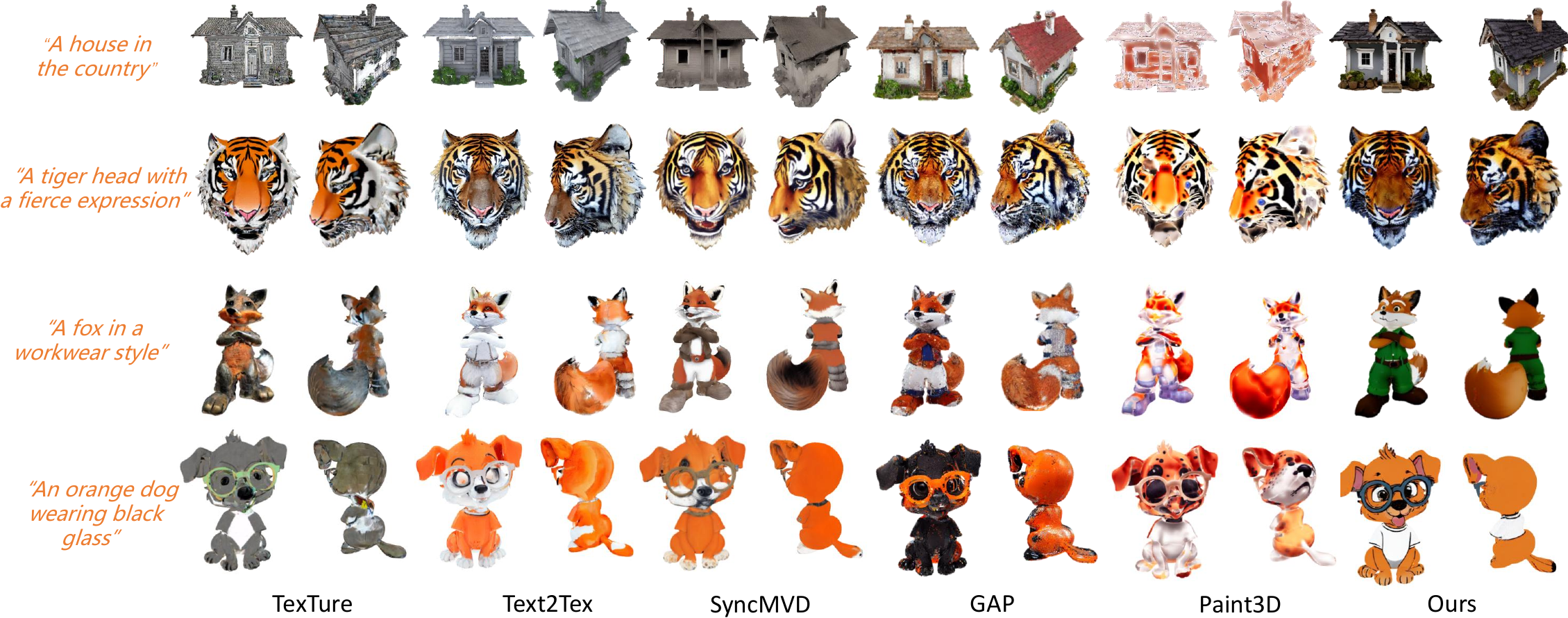}
    \vspace{-0.8cm}
    \caption{Visual comparison on the Objaverse dataset shows that GaussianGrow uses point clouds instead of meshes.}
    \label{fig:quality_objaverse}
    \vspace{-0.8cm}
    
\end{figure*}

\noindent\textbf{Image-level Inpainting.}
Once we have identified the optimal camera position $v_i$, which observes the largest unseen region, we need to generate a complete and high-fidelity image from the new viewpoint as the supervision for Gaussian inpainting. Specifically, we first detect unoptimized Gaussians in 2D space to create a generated mask $M$ as the inpainting condition. The generated mask $M$, along with the depth map $D_i$ rendered from $v_i$, the rendered occluded image $I_i$, and text prompt $c$, is fed into a depth-aware inpainting diffusion model based on Stable Diffusion~\cite{rombach2021highresolution} and ControlNet~\cite{zhang2023adding} to inpaint $I_i$ into a high-quality complete image $\hat{I}_i$.

\noindent\textbf{Iterative Gaussian Inpainting.}
Given the inpainted image $\hat{I}_i$ at the optimal camera position $v_i$, which observes the largest unseen region at this step, we optimize the invisible Gaussians with supervisions from the inpainted parts of $\hat{I}_i$, using the same strategies in Sec.~\ref{sec.3.2}. Since the unseen regions are inpainted in the current view, we iteratively perform Gaussian inpainting by identifying the camera poses that observe the largest remaining unseen regions in the next step. This iteration continues until all Gaussians are generated, where we empirically find that six iterations are sufficient for most models.
The key advantage of our approach lies in its adaptability to complex geometries, as it dynamically identifies and prioritizes occluded regions instead of relying on predefined viewpoint patterns, achieving more complete coverage with fewer views. We show a Gaussian generation before and after our Gaussian inpainting in Fig.~\ref{fig:ablation}.



\noindent\textbf{Spatial Inpainting.}
Due to noises and uneven density in the raw point cloud data, some points may remain difficult to observe after image inpainting-based Gaussian inpainting. To address this, we adopt a spatial inpainting approach as a post-processing step to complete the Gaussians. The process implements a propagation mechanism that transfers properties from optimized neighbors to nearby unoptimized Gaussians.




\section{Experiments}
\label{sec:exp}
In this section, we present a comprehensive evaluation of GaussianGrow's performance across multiple scenarios. We begin by assessing GaussianGrow's capabilities in text-guided visual synthesis in Sec.~\ref{sec:4.1}. We then conduct experiments on text-to-3D generation in Sec.~\ref{sec.t23d}. In Sec.~\ref{sec:4.2}, we make a comparison with existing approaches for point-to-Gaussian generation. Finally, we validate our design choices through detailed ablation studies in Sec.~\ref{sec:4.4}.

\begin{table}[t!]
   \caption{
    Quantitative comparison on the Objaverse dataset.
   } 
   \vspace{-0.4cm}
   \label{tab:quantitative_objaverse}
   \centering
   \resizebox{\linewidth}{!}{
       \begin{tabular}{l|ccc|cc}
       \toprule
        \textbf{Method} & \textbf{FID$\downarrow$} & \textbf{KID$\downarrow$} & {\textbf{CLIP$\uparrow$}} & \multicolumn{2}{c}{\textbf{User Study} } \\
        & & & & \textbf{Overall Quality$\uparrow$} & \textbf{Text Fidelity$\uparrow$} \\ 
       \midrule
        \textbf{TexTure} \cite{richardson2023texture}  & 42.63 & 7.84 & 26.84 & 1.49 & 1.67
\\ 
        \textbf{Text2Tex} \cite{chen2023text2tex} & 41.62 & 6.45 & 26.73 & 2.37 & 3.23
\\
        \textbf{SyncMVD} \cite{liu2024text}  & 40.85 & 5.77 & 27.24 & 4.13 & 4.34\\ 
        \textbf{Paint3D} \cite{zeng2024paint3d}  & 41.08 & 5.81 & 26.73 & 3.37 & 3.45\\
        \textbf{GAP} \cite{gap}  & 40.39 & 5.28 & 27.26 & 3.37 & 4.13\\ 
        \midrule
        \textbf{TexTure}\textsubscript{\textit{BPA}} & 60.69 & 15.98 & 26.62 & \textemdash{} & \textemdash{} \\ 
        \textbf{Text2Tex}\textsubscript{\textit{BPA}} & 64.35 & 16.67 & 26.18 & \textemdash{} & \textemdash{} \\
        \textbf{SyncMVD}\textsubscript{\textit{BPA}}  & 60.29 & 14.35 & 26.19 & \textemdash{} & \textemdash{}\\ 
        \textbf{Paint3D}\textsubscript{\textit{BPA}}  & 65.36 & 17.37 & 25.14 & \textemdash{} & \textemdash{}\\ 
        \midrule
        \textbf{TexTure}\textsubscript{\textit{CAP}}  & 53.55 & 12.43 & 26.68 & \textemdash{} & \textemdash{}\\ 
        \textbf{Text2Tex}\textsubscript{\textit{CAP}} & 52.78 & 11.09 & 26.78 & \textemdash{}& \textemdash{}\\
        \textbf{SyncMVD}\textsubscript{\textit{CAP}}  & 63.85 & 16.92 & 25.81 & \textemdash{}& \textemdash{} \\ 
        \textbf{Paint3D}\textsubscript{\textit{CAP}}  & 59.49 & 13.56 & 24.99 & \textemdash{} & \textemdash{}\\ 
        \midrule
        \cellcolor{mygray}{\textbf{Ours}}  & \cellcolor{mygray}{\textbf{36.07}}& \cellcolor{mygray}{\textbf{3.04}}& \cellcolor{mygray}{\textbf{27.30}} & \cellcolor{mygray}{\textbf{4.67}} &  \cellcolor{mygray}{\textbf{4.72}}\\  
        \bottomrule
       \end{tabular}
   }
       \vspace{-0.8cm}
\end{table}

\subsection{Text-Guided Visual Synthesis}
\label{sec:4.1}
\noindent\textbf{Datasets and Evaluation Metrics.} In line with previous work~\cite{chen2023text2tex, richardson2023texture}, our experiments leverage a curated collection from the Objaverse database~\cite{objaverse}, encompassing 410 detailed 3D models across 225 distinct categories. Unlike many competing approaches that require complete mesh representations, GaussianGrow operates directly on point cloud inputs, without additional geometric information.

\begin{table*}[t]
\setlength{\tabcolsep}{4pt}
    \centering
    \caption{Quantitative evaluations on T3Bench benchmark for text-to-3D generation. Higher is better for all metrics.}
    \vspace{-0.4cm}
    \label{tab:t3bench}
    \renewcommand\arraystretch{1.5}
    \resizebox{\textwidth}{!}{
        \begin{tabular}{l|c|c|c|c|c|c|c|c|c}
            \toprule[1.2pt]
 ~ & \cellcolor{mygray}{\ \ \ \ \ \ Ours + Uni3D\ \ \ \ \ \ } & Ours + LGM & DiffSplat \cite{lin2025diffsplat} & GVGEN \cite{he2024gvgen} & LN3Diff \cite{lan2024ln3diff} & DIRECT-3D \cite{liu2024direct} & 3DTopia \cite{hong20243dtopia} & LGM \cite{tang2024lgm} & GRM \cite{xu2024grm} \\

            \midrule[1.2pt]

            $\uparrow$ CLIP Sim.$_\%$ & 
            \cellcolor{mygray}{\textbf{31.55}} & 30.17 & 30.95 & 23.66 & 24.36 & 24.80 & 25.55 & 29.96 & 28.19 \\
            $\uparrow$ CLIP R-Prec.$_\%$ &
            \cellcolor{mygray}{\textbf{82.00}} & 81.00 & 81.00 & 23.25 & 27.25 & 30.75 & 34.50 & 78.00 & 64.75 \\
            $\uparrow$ ImageReward &
            \cellcolor{mygray}{\textbf{-0.316}} & -0.329 & -0.491 & -2.156 & -2.008 & -2.005 & -1.998 & -0.720 & -1.337 \\
            
            \bottomrule[1.2pt]
            
        \end{tabular}
    }
    \vspace{-0.6cm}
\end{table*}

\begin{figure}[b]
    \centering
    \vspace{-0.4cm}
    \includegraphics[width=1\linewidth]{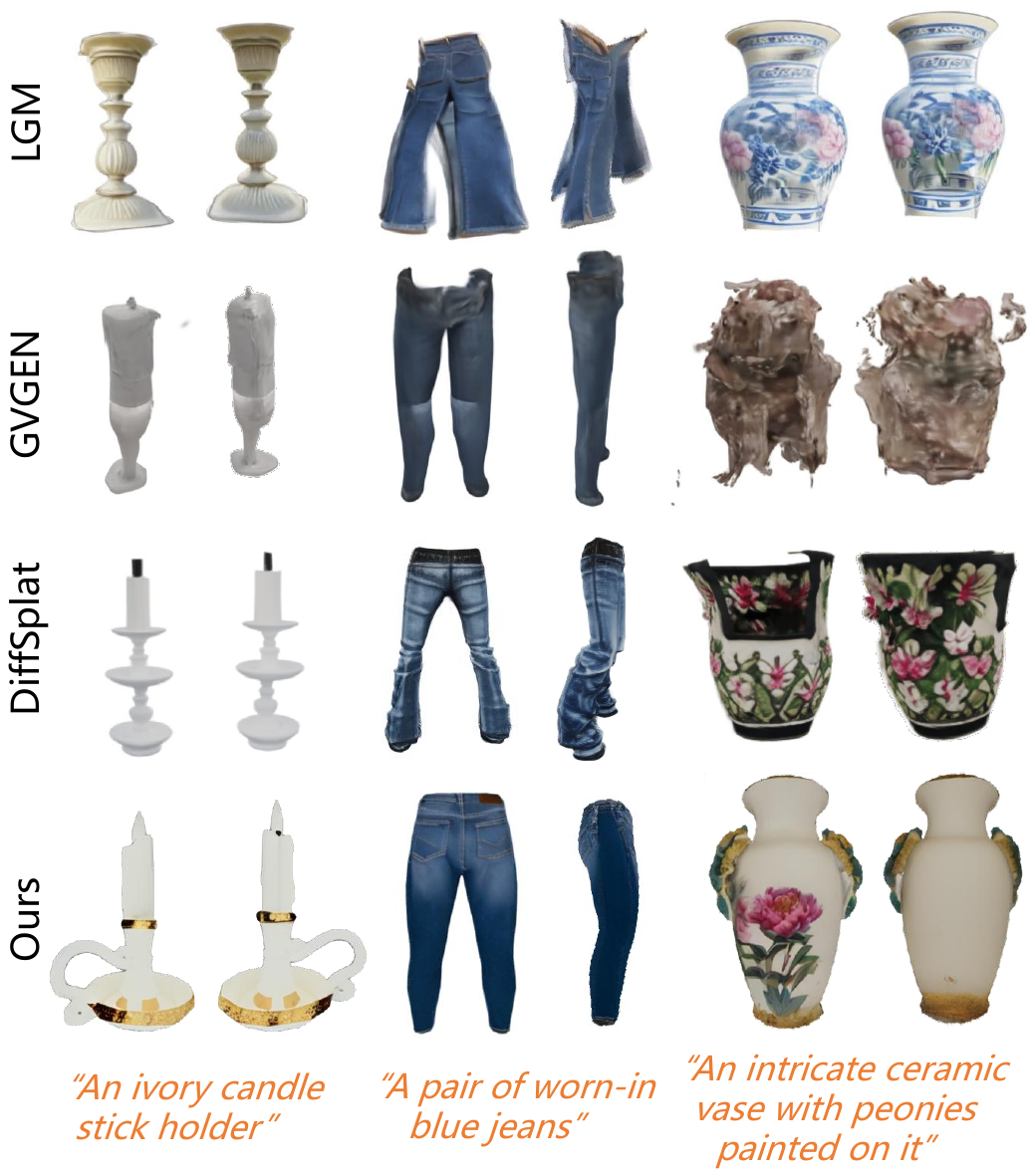}
    \vspace{-0.6cm}
    \caption{Text-to-3D comparisons on T3Bench.}
    \label{fig.t3bench}
    \vspace{-0.6cm}
    
\end{figure}

For quantitative evaluation, we employ three complementary metrics: Fréchet Inception Distance (FID)~\cite{Kynkaanniemi2022} and Kernel Inception Distance (KID $ \times 10^{-3}$)~\cite{binkowski2018demystifying} to assess image quality, while the alignment between generated content and textual prompts is measured using CLIP Score~\cite{radford2021learning}. All evaluations are performed on high resolution $1024 \times 1024$ renderings captured from standardized viewpoints to ensure fair comparisons.

\noindent\textbf{Baselines and Implementations.} We benchmark GaussianGrow against state-of-the-art text-guided 3D appearance generation methods, including Texture~\cite{richardson2023texture}, Text2Tex~\cite{chen2023text2tex}, Paint3D~\cite{zeng2024paint3d}, SyncMVD~\cite{liu2024text} and GAP~\cite{gap}. 
Unlike most of these methods, which rely on UV-mapped meshes, GaussianGrow operates directly on point clouds. To ensure a fair comparison, we further report performance of the baseline methods with reconstructed meshes from point clouds using the Ball-Pivoting Algorithm (BPA)\cite{bernardini1999ball} and CAP-UDF\cite{zhou2022capudf}, followed by UV map generation via xatlas unwrapping~\cite{xatlas}.

\noindent\textbf{Comparison.} Our quantitative evaluation demonstrates that GaussianGrow consistently outperforms existing state-of-the-art methods, as shown in Tab.~\ref{tab:quantitative_objaverse}. 
The performance gap becomes even more significant when comparing against baseline methods operating on reconstructed meshes. The mesh reconstruction process introduces several challenges: geometric details often get smoothed or distorted during point-to-mesh conversion, and topological errors can emerge in complex regions. 
In contrast, GaussianGrow bypasses UV parameterization entirely by directly optimizing Gaussian primitives in 3D space. Visual comparisons in Fig.~\ref{fig:quality_objaverse} demonstrate that GaussianGrow better preserves fine details and achieves more consistent texturing across complex surfaces, particularly in geometrically intricate regions.

\begin{figure*}[ht]
    \centering
    \includegraphics[width=0.9\linewidth]{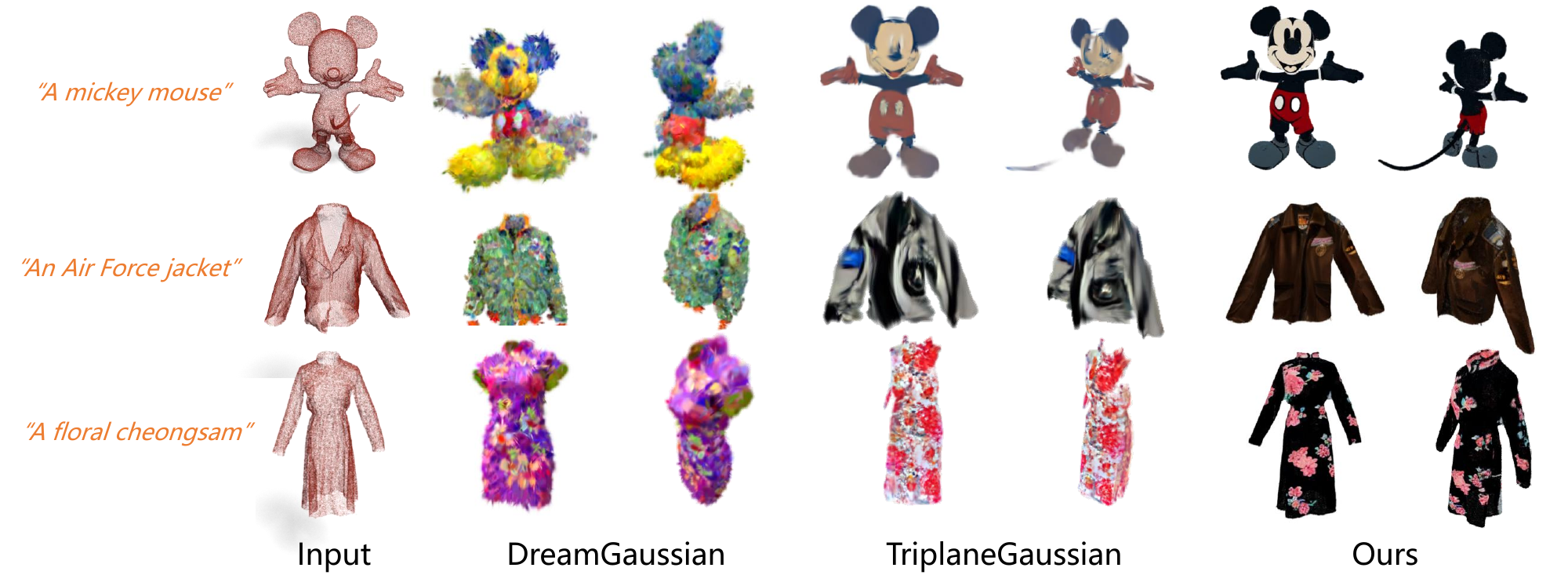}
    \vspace{-0.4cm}
    \caption{Visual comparison with DreamGaussian and TriplaneGaussian on the task of Point-to-Gaussian.}
    \label{fig:point2gs}
    \vspace{-0.5cm}
\end{figure*}

\subsection{Text-to-3D Generation}
\label{sec.t23d}

\noindent\textbf{Implementation and Benchmark.} GaussianGrow enables text-to-3D generation with two geometric settings including the retrieve-based setting and the generative-based setting. For the retrieve-based setting, we employ Uni3D \cite{zhou2023uni3d} to retrieve reference point clouds from the G-Objaverse dataset \cite{qiu2024richdreamer}, a carefully curated subset of Objaverse \cite{objaverse}, based on the input text prompt. The retrieved point clouds serve as geometric priors that guide the generation process. For a generative-based setting, we employ LGM~\cite{tang2024lgm} for generating scaffold point clouds as the geometric prior. GaussianGrow then generates 3D Gaussians conditioned on both the retrieved or generated point clouds and the corresponding text description, enabling the model to leverage both geometric structure and semantic information for high-quality synthesis. The results for the two geometric settings are reported as ``Ours+Uni3D'' and ``Ours+LGM''.

For quantitative evaluation, we conduct comprehensive experiments on the T3Bench benchmark \cite{he2023t}, which provides a diverse collection of text prompts covering various object categories and complexity levels. We measure performance using three complementary metrics: CLIP similarity for semantic alignment, CLIP R-Precision for text-image correspondence, and ImageReward \cite{radford2021learning, xu2023imagereward} for perceptual quality assessment.

\noindent\textbf{Comparison.} We benchmark our method against state-of-the-art text-to-3D approaches including DiffSplat \cite{lin2025diffsplat}, GVGEN \cite{he2024gvgen}, LN3Diff \cite{lan2024ln3diff}, DIRECT-3D \cite{liu2024direct}, 3D-Topia \cite{hong20243dtopia}, LGM \cite{tang2024lgm} and GRM \cite{xu2024grm}. Visual results in Fig.~\ref{fig.t3bench} further demonstrate that our method produces more realistic and text-aligned 3D content.

The retrieve-based GaussianGrow ``Ours+Uni3D'' achieves the best performance across all evaluation metrics, while the generative-based version ``Ours+LGM'' also achieves comparable performance compared to the state-of-the-art method DiffSplat. Moreover, applying the geometry of LGM to GaussianGrow also achieves significantly better performance by replacing the appearance of LGM with GaussianGrow. The results demonstrate that GaussianGrow significantly outperforms previous methods in terms of appearance generation, and a stronger geometric setting leads to better generation quality. 


\subsection{Point to Gaussian Generation}
\label{sec:4.2}
\noindent\textbf{Datasets.} To evaluate GaussianGrow's effectiveness in Point-to-Gaussian generation, we experimented with two representative datasets. For Objaverse, we sampled $100K$ points from each mesh to create synthetic point clouds with high geometric fidelity. To demonstrate robustness with real-world data, we also utilized the DeepFashion3D dataset containing real-scanned point clouds. These scans present challenging characteristics including noise and varying point densities.

\noindent\textbf{Baselines and Implementations.} We benchmark GaussianGrow against two leading methods: DreamGaussian~\cite{tang2023dreamgaussian} and TriplaneGaussian~\cite{zou2023triplane}. Each baseline required specific adaptations for our experiments. We modified DreamGaussian by replacing its random initialization with point cloud guidance, enabling direct point input. TriplaneGaussian was adapted by bypassing its point cloud decoder for direct point-to-Gaussian conversion and integrating Stable Diffusion for text guidance.

\noindent\textbf{Comparison.} As shown in Fig.~\ref{fig:point2gs}, our visual comparisons highlight that GaussianGrow delivers noticeably better visual quality and geometric fidelity than baseline methods. DreamGaussian, while utilizing SDS~\cite{poole2022dreamfusion} for appearance optimization, often yields oversaturated and unnatural colors and requires costly, parameter-sensitive optimization. TriplaneGaussian~\cite{zou2023triplane} is constrained by the limited resolution of its triplanes, hindering its ability to recover fine-grained appearance and complex geometry.

\begin{figure}
    \centering
    \includegraphics[width=1\linewidth]{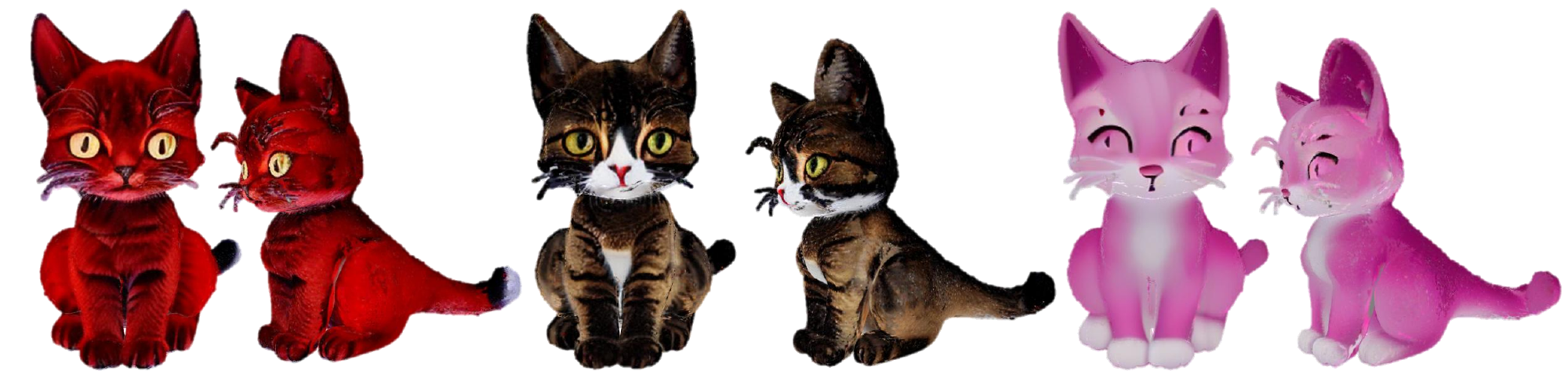}
    \vspace{-0.6cm}
    \caption{Diverse style Gaussian generations.}
    \label{fig:diversity}
    \vspace{-0.8cm}
\end{figure}

\subsection{Ablation Study}
\label{sec:4.4}
We evaluate our key components through ablation experiments. Fig. ~\ref{fig:ablation} shows that the removal of our image-level inpainting strategy leaves complex regions incomplete. The visual results clearly show the effectiveness of our inpainting mechanism in generating coherent appearances for areas with limited visibility.

To quantitatively evaluate the contribution of each component, we conducted ablation studies focused on two key modules: our Overlap Detection with Camera Pose Optimization strategy and the Image-level Inpainting process. Tab.~\ref{tab:ablation} presents the results across our evaluation metrics. 

To assess the effectiveness of our overlap-region strategy, we examine the influence of the number of generation views $K$. As shown in Tab.~\ref{tab:hyperparameter}, using only the six cardinal views leads to clear degradation across all metrics, while adding four views focused on key overlap regions yields the best performance. Increasing $K$ beyond this offers negligible gains, indicating that our camera-pose optimization already captures the most critical overlapping areas and that additional views add cost with little benefit.

\noindent\textbf{Diverse Style Gaussians.} GaussianGrow effectively generates varied appearances for identical geometric inputs by simply changing text prompts. As demonstrated in Fig.~\ref{fig:diversity}, the same point cloud processed with different textual descriptions produces distinct visual styles while maintaining geometric accuracy.

\begin{table}[t]
\vspace{-0.2cm}
   \caption{
    Ablation results for key components of GaussianGrow.
   } 
   \vspace{-0.3cm}
    \label{tab:ablation}
    \small
    \centering
    \setlength{\tabcolsep}{1.2em}
    \renewcommand{\arraystretch}{1.0}
    \begin{tabularx}{\linewidth}{>{\centering}m{3.5cm}| Y Y Y}
    \toprule
    \textbf{Method} & \textbf{FID$\downarrow$} & \textbf{KID$\downarrow$} & {\textbf{CLIP$\uparrow$}} \\
    \midrule
    \cellcolor{mygray}{\textbf{Full Model}}  & \cellcolor{mygray}{\textbf{36.07}} & \cellcolor{mygray}{\textbf{3.04}} & \cellcolor{mygray}{\textbf{27.30}} \\ 
    \midrule
    \textbf{W/o Overlap Processing}  & 40.48 & 4.81 & 26.73 \\ 
    \textbf{W/o Inpaint}  & 40.46 & 4.68 & 26.71 \\ 
    \bottomrule
    \end{tabularx}
    \vspace{-0.8cm}
\end{table}

\begin{table}[h]
\vspace{-0.3cm}
\centering
\caption{Impact of the number of views $K$ on generation quality.}
\label{tab:hyperparameter}
\resizebox{0.45\textwidth}{!}{
\begin{tabular}{l|ccc}
\hline
Number of Views & FID$\downarrow$ & KID$\downarrow$ & CLIP$\uparrow$ \\
\hline
$K = 6$ (cardinal only) & 40.48 & 4.81 & 26.73 \\
$K = 10$ & \textbf{36.07} & \ 3.04 & \textbf{27.30} \\
$K = 12$ & 36.57 & \textbf{2.88} & 26.48 \\
\hline
\end{tabular}}
\vspace{-0.6cm}
\end{table}

\section{Conclusion}

We introduce GaussianGrow, a novel approach for generating 3D Gaussians by growing them from readily available point clouds. Our method leverages a text-guided multi-view diffusion model for appearance synthesis while constraining novel views to reduce fusion artifacts. Additionally, we iteratively complete hard-to-observe regions via pose-aware inpainting. Experiments on synthetic and real-scanned point clouds demonstrate the effectiveness of GaussianGrow in generating high-fidelity 3D Gaussians.

\label{sec:con}

\section{Acknowledgment}
This work was supported by Deep Earth Probe and Mineral Resources Exploration - National Science and Technology Major Project (2024ZD1003405), and the National Natural Science Foundation of China (62272263), and in part by Kuaishou. Junsheng Zhou is also partially funded by Baidu Scholarship.


{
    \small
    \bibliographystyle{ieeenat_fullname}
    \bibliography{main}
}

\end{document}